\documentclass[letterpaper]{article} % DO NOT CHANGE THIS

\usepackage{subcaption}
\usepackage{aaai2026}  % DO NOT CHANGE THIS
\usepackage{times}  % DO NOT CHANGE THIS
\usepackage{helvet}  % DO NOT CHANGE THIS
\usepackage{courier}  % DO NOT CHANGE THIS
\usepackage[hyphens]{url}  % DO NOT CHANGE THIS
\usepackage{graphicx} % DO NOT CHANGE THIS
\usepackage{natbib}  % DO NOT CHANGE THIS AND DO NOT ADD ANY OPTIONS TO IT
\usepackage{caption} % DO NOT CHANGE THIS AND DO NOT ADD ANY OPTIONS TO IT
\usepackage{algorithm}
\usepackage{algorithmic}
\usepackage{amsmath}
\usepackage{pifont}
\newcommand{\cmark}{\ding{51}}
\newcommand{\xmark}{\ding{55}}
\usepackage{newfloat}
\usepackage{listings}
\DeclareCaptionStyle{ruled}{labelfont=normalfont,labelsep=colon,strut=off} % DO NOT CHANGE THIS
\floatstyle{ruled}
\newfloat{listing}{tb}{lst}{}
\floatname{listing}{Listing}
\title{LISA: A Layer-wise Integration and Suppression Approach for Hallucination Mitigation in Multimodal Large Language Models}
\author{
    Zhihui Guo\textsuperscript{\rm 1},
    Xin Man\textsuperscript{\rm 1},
    Hui Xu\textsuperscript{\rm 1},
    Jie Shao\textsuperscript{\rm 1,2}
    Zhiguo Jiang\textsuperscript{\rm 3},
    Xianchao Zhang\textsuperscript{\rm 3},
    Heng Tao Shen\textsuperscript{\rm 2}
    }
\affiliations{
    \textsuperscript{\rm 1}Shenzhen Institute for Advanced Study, University of Electronic Science and Technology of China\\
    \textsuperscript{\rm 2}Sichuan Artificial Intelligence Research Institute, Yibin\\
    \textsuperscript{\rm 3}Provincial Key Laboratory of Multimodal Perceiving and Intelligent Systems, Jiaxing University\\
    \{zhihuiguo,manxin\}@std.uestc.edu.cn, \{huixu.kim,shaojie\}@uestc.edu.cn, xidianzhiguo@163.com, zhangxianchao@zjxu.edu.cn, shenhengtao@hotmail.com}

\usepackage{bibentry}
\begin{document}

\maketitle

\begin{abstract}
Multimodal Large Language Models (MLLMs) excel in vision-language
tasks such as image captioning but remain prone to object
hallucinations, where they describe objects that do not appear in
the image. To mitigate this, we propose \textbf{LISA}, a
\textbf{L}ayer-wise \textbf{I}ntegration and \textbf{S}uppression
\textbf{A}pproach. LISA leverages the layer-wise functional roles in
MLLMs: shallow layers provide visual grounding, middle layers encode
semantics, and deep layers tend to amplify spurious signals. First,
layer-wise spectral modulation stabilizes attention by suppressing
over-amplified activations in deeper layers while preserving
alignment cues in earlier layers. Second, token-level logits from
selected layers are fused via anchor-based routing, with token-wise
anchor selection and soft logit fusion enabling adaptive integration
during decoding. LISA is fully \textbf{plug-and-play} and can be
seamlessly integrated into existing MLLMs, including Qwen2.5-VL.
Experiments on multiple benchmarks show that LISA reduces
hallucinations by up to 53.6\% in $\text{CHAIR}_\text{I}$ and
improves POPE F1 by up to 5.1\%, demonstrating strong generalization
across models and tasks. Our code is available at
\url{https://github.com/zhlisa1010-eng/LISA}.
\end{abstract}

\section{Introduction}

\begin{figure}[t]
    \centering
    \includegraphics[width=0.95\linewidth]{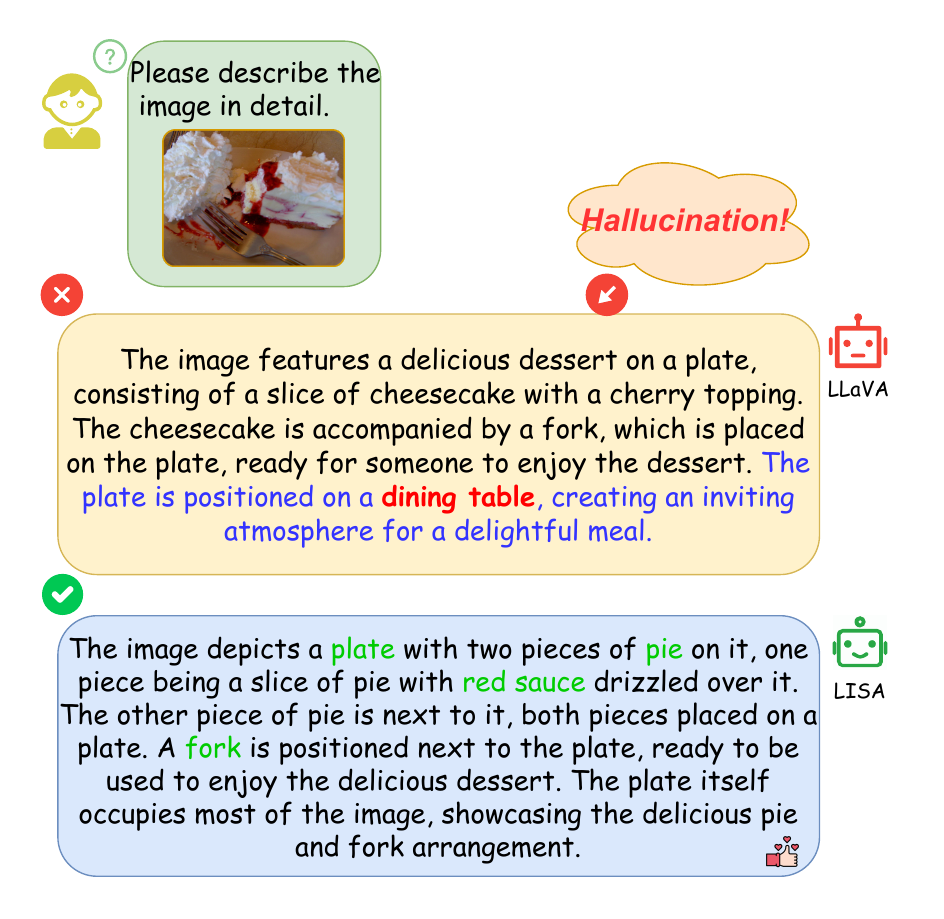}
    \caption{\textbf{An illustration of \textbf{object hallucination} in
multimodal generation.} The upper response, generated by the LLaVA
model \cite{DBLP:conf/cvpr/LiuLLL24} using greedy decoding,
hallucinates non-existent objects---``dining'' and ``table''
(highlighted in red)---which are absent from the image. This reveals
a typical failure of visual grounding. In contrast, the lower
response, produced by LISA, provides a faithful and grounded
description, free from \textbf{object hallucination}.}
    \label{fig:hallucination_lisa}
\end{figure}

The rapid progress of Large Language Models (LLMs)
\cite{DBLP:journals/mima/FloridiC20,DBLP:journals/corr/abs-2302-13971,DBLP:journals/corr/abs-2303-08774,DBLP:journals/corr/abs-2502-06604}
has fostered Multimodal Large Language Models (MLLMs)
\cite{DBLP:conf/cvpr/LiuLLL24,DBLP:conf/nips/Dai0LTZW0FH23,DBLP:journals/corr/abs-2308-12966},
which unify vision-language tasks such as image captioning and
visual question answering. Despite these advancements, a persistent
issue remains: MLLMs frequently exhibit \textbf{object
hallucination}
\cite{DBLP:conf/emnlp/RohrbachHBDS18,DBLP:conf/mm/ZhangXSHCKWW25}, a
specific type of hallucination where the model describes concrete
objects that are not present in the image. As illustrated in
Figure~\ref{fig:hallucination_lisa}, the model incorrectly mentions
a ``dining'' and a ``table'' that do not appear in the visual input,
revealing a typical failure of visual grounding.

To tackle this challenge, existing studies have explored a range of
solutions, including training-based
\cite{DBLP:conf/cvpr/YangLHX025,DBLP:conf/iclr/LiuLLWYW24,DBLP:conf/mm/ShangZZYFZCLT25,DBLP:conf/mm/DaiLZWL25,DBLP:conf/mm/ZhaoZLXYTFCWZ25}
and decoding-based
\cite{DBLP:conf/aaai/WuFPWYC25,DBLP:journals/chinaf/YinFZXWSSLSC24,DBLP:conf/iclr/ChuangXLKGH24,DBLP:conf/iclr/TangHLSYL25,DBLP:conf/cvpr/LengZCLLMB24,DBLP:conf/cvpr/HuangDZ0H0L0Y24,DBLP:conf/iclr/Wang00TXDC25,DBLP:conf/mm/CaoHLXWC25,DBLP:conf/mm/XieGYXXKFQYGMGYWZ25,DBLP:conf/mm/HuDZK25,DBLP:conf/mm/LiWJCJSCJ25,DBLP:conf/mm/RuanZGYLF25}
approaches. Training-based methods typically optimize model
parameters using hallucination-sensitive supervision. OPA-DPO
\cite{DBLP:conf/cvpr/YangLHX025}, for instance, distills visual
preference signals from the closed-source GPT-4V to guide model
responses toward human-aligned outputs, but this reliance on
proprietary systems limits reproducibility and transparency. In
contrast, LRV-Instruction \cite{DBLP:conf/iclr/LiuLLWYW24} curates a
large-scale instruction dataset comprising both grounded and
hallucination-inducing prompts to enhance robustness via instruction
tuning. Despite avoiding manual annotation, these approaches require
substantial effort for dataset construction and model adaptation,
constraining their scalability.

Given the high cost and limited scalability of training-based
methods, decoding-time strategies have emerged as scalable,
training-free alternatives for hallucination mitigation. Some
methods leverage external tools to enhance reasoning or
verification. Bottom-Up Holistic Reasoning
\cite{DBLP:conf/aaai/WuFPWYC25} uses external tools such as scene
graph parsers, object detectors, and web search to perform
structured reasoning during decoding, while Woodpecker
\cite{DBLP:journals/chinaf/YinFZXWSSLSC24} performs post-hoc
multi-stage analysis to detect and correct hallucinated outputs.
Other approaches focus on internal model behavior. DoLa
\cite{DBLP:conf/iclr/ChuangXLKGH24} contrasts token confidence
across layers to suppress unsupported outputs; TAME
\cite{DBLP:conf/iclr/TangHLSYL25} regularizes the query-key
eigenspectrum to stabilize signal propagation; VCD
\cite{DBLP:conf/cvpr/LengZCLLMB24} applies contrastive constraints
on matched and mismatched visual regions; OPERA
\cite{DBLP:conf/cvpr/HuangDZ0H0L0Y24} re-decodes with penalties to
discourage ungrounded content; and DeCo
\cite{DBLP:conf/iclr/Wang00TXDC25} fuses intermediate-layer logits
to guide more grounded generation. However, most existing decoding
methods treat all transformer layers as equally informative
\cite{DBLP:conf/iclr/HuangZSCZXYL25,DBLP:conf/iclr/ChoKHC25,DBLP:journals/tmm/RuTXLS24,DBLP:conf/iclr/Fuh0SHQN25,DBLP:conf/cvpr/000100N00H25},
without considering their functional differences. Based on empirical
analysis of model behaviors across layers, we observe that MLLMs
exhibit internal functional hierarchy: shallow layers contribute to
visual grounding, middle layers encode semantic context, and deep
layers tend to amplify unstable signals. These observations
underscore the need for decoding strategies that explicitly account
for layer-wise functional roles and modulate information across
layers accordingly.

Building on this observation, we propose \textbf{LISA} (\textit{a
\textbf{L}ayer-wise \textbf{I}ntegration and \textbf{S}uppression
\textbf{A}pproach}), a training-free decoding strategy designed to
mitigate object hallucination. LISA exploits layer-wise functional
roles in multimodal transformers, partitioning the model into
shallow, middle, and deep zones for grounding, semantic integration,
and spurious-signal suppression. It stabilizes the attention weights
via \textit{layer-wise spectral modulation}, suppressing
over-amplified, unstable signals in deeper layers while preserving
stable alignment cues in shallower ones. Token-level logits from
representative layers are then aggregated through cross-layer fusion
and token-wise anchor routing, ensuring more reliable and faithful
generation.

LISA is fully \textbf{plug-and-play}, requiring no retraining or
architectural modification, and can be seamlessly applied to a
variety of MLLMs. We evaluate LISA on several hallucination
benchmarks, including CHAIR \cite{DBLP:conf/emnlp/RohrbachHBDS18},
POPE \cite{DBLP:conf/emnlp/LiDZWZW23}, AMBER
\cite{DBLP:journals/corr/abs-2311-07397}, and MME
\cite{DBLP:journals/corr/abs-2306-13394}, demonstrating consistent
hallucination reduction and improved factual consistency. Our method
offers a lightweight yet effective approach to enhance the
reliability of multimodal generation.

Our contributions are summarized as follows:
\begin{itemize}
    \item We identify a key limitation of existing decoding-time approaches:
they overlook the layer-wise functional roles of transformer layers,
leading to suboptimal integration of layer-wise representations and
insufficient modulation of spectral instability during decoding.
    \item We propose LISA, a training-free decoding framework that leverages
the functional hierarchy of MLLMs. By applying layer-wise spectral
modulation to stabilize attention dynamics and fusing logits across
selected anchor layers, LISA effectively mitigates hallucinations.
    \item We conduct comprehensive evaluations on four hallucination
benchmarks across multiple MLLMs. LISA consistently reduces
hallucinations while preserving generation quality and demonstrates
strong compatibility with diverse architectures and decoding
strategies.
\end{itemize}

\section{Related Work}

\subsection{Hallucination in MLLMs}

Multimodal Large Language Models (MLLMs) unify visual and linguistic
understanding to handle tasks such as image captioning and visual
question answering. Despite advances in instruction tuning and
cross-modal alignment
\cite{DBLP:conf/cvpr/LiuLLL24,DBLP:conf/nips/Dai0LTZW0FH23,DBLP:journals/corr/abs-2308-12966,DBLP:journals/corr/abs-2501-12327,DBLP:journals/corr/abs-2505-20897,DBLP:conf/acl/YinXYYRZLZ25,DBLP:journals/corr/abs-2504-02949,DBLP:journals/tmm/LiXMSPS25},
MLLMs still frequently hallucinate object-level details that are
unsupported by the input image, especially in complex or ambiguous
scenes. These failures expose fundamental limitations in visual
grounding and continue to drive research on hallucination
mitigation.

\subsection{Hallucination Mitigation for MLLMs}

Decoding-time strategies have received increasing attention for
their training-free and plug-and-play characteristics. Methods such
as DoLa \cite{DBLP:conf/iclr/ChuangXLKGH24}, DeCo
\cite{DBLP:conf/iclr/Wang00TXDC25}, OPERA
\cite{DBLP:conf/cvpr/HuangDZ0H0L0Y24}, and VCD
\cite{DBLP:conf/cvpr/LengZCLLMB24} intervene during generation by
reweighting logits, anchoring shallow-layer signals, or reranking
outputs. While effective, these approaches often ignore the
functional hierarchy of transformer layers, treating them as
uniformly informative.

In contrast, our method LISA introduces a layer-wise perspective on
the transformer architecture. Instead of treating all layers
equivalently, LISA explicitly partitions the model into shallow,
middle, and deep zones based on their observed roles in grounding
visual entities, maintaining contextual coherence, and refining
linguistic details. To stabilize generation, LISA performs targeted
spectral modulation across these zones, dynamically suppressing
unstable signals while preserving grounded information. This
layer-wise intervention enables more precise control over the
generation process and improves alignment between textual outputs
and visual evidence.

\subsection{Layer-wise Structure in MLLMs}

\citet{DBLP:conf/cvpr/0009YHS25} conducted attention ablation
analyses, revealing a staged organization within MLLMs: shallow
layers propagate general visual features, intermediate layers refine
object-specific details, and later layers consolidate task-specific
semantics. While their investigation offers valuable
interpretability, it remains primarily descriptive, abstaining from
a systematic exploration of how to leverage this hierarchical
structure to enhance generative capabilities.

In contrast, our proposed LISA builds on this insight by explicitly
treating the transformer as a functionally hierarchical
system---dividing it into shallow, middle, and deep zones---and
actively stabilizing cross-layer representations through layer-wise
spectral suppression and cross-layer fusion. This design transforms
descriptive observations into an effective, plug-and-play decoding
mechanism that improves generation robustness without requiring
retraining or architectural changes.

\begin{figure*}[t]
  \centering
  \captionsetup[subfigure]{labelformat=empty}
  \begin{subfigure}{0.3\textwidth}
    \centering
    \includegraphics[width=\linewidth]{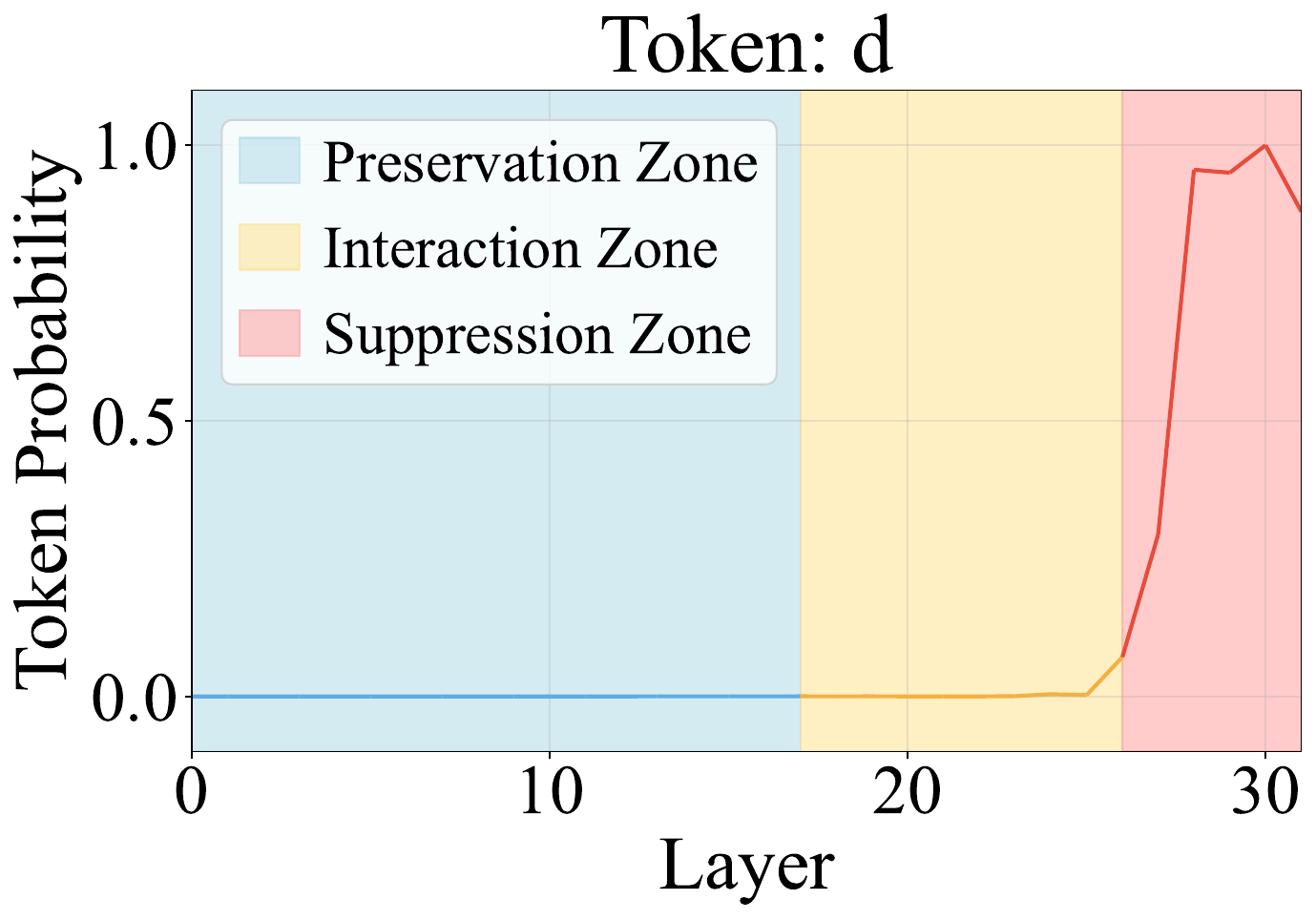}
    \subcaption{(a\textsubscript{1})}
  \end{subfigure}
  \hfill
  \begin{subfigure}{0.3\textwidth}
    \centering
    \includegraphics[width=\linewidth]{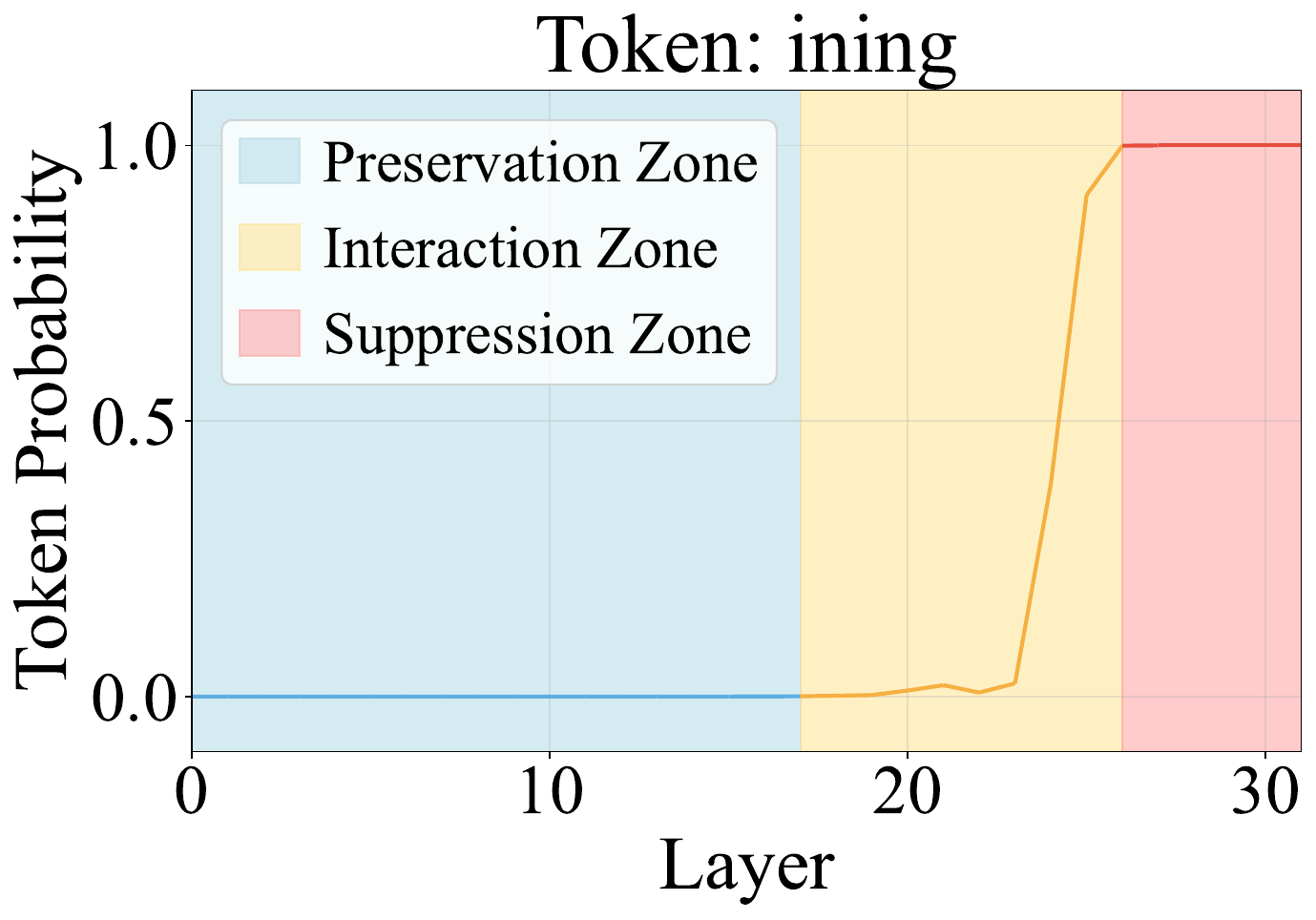}
    \subcaption{(a\textsubscript{2})}
  \end{subfigure}
  \hfill
  \begin{subfigure}{0.3\textwidth}
    \centering
    \includegraphics[width=\linewidth]{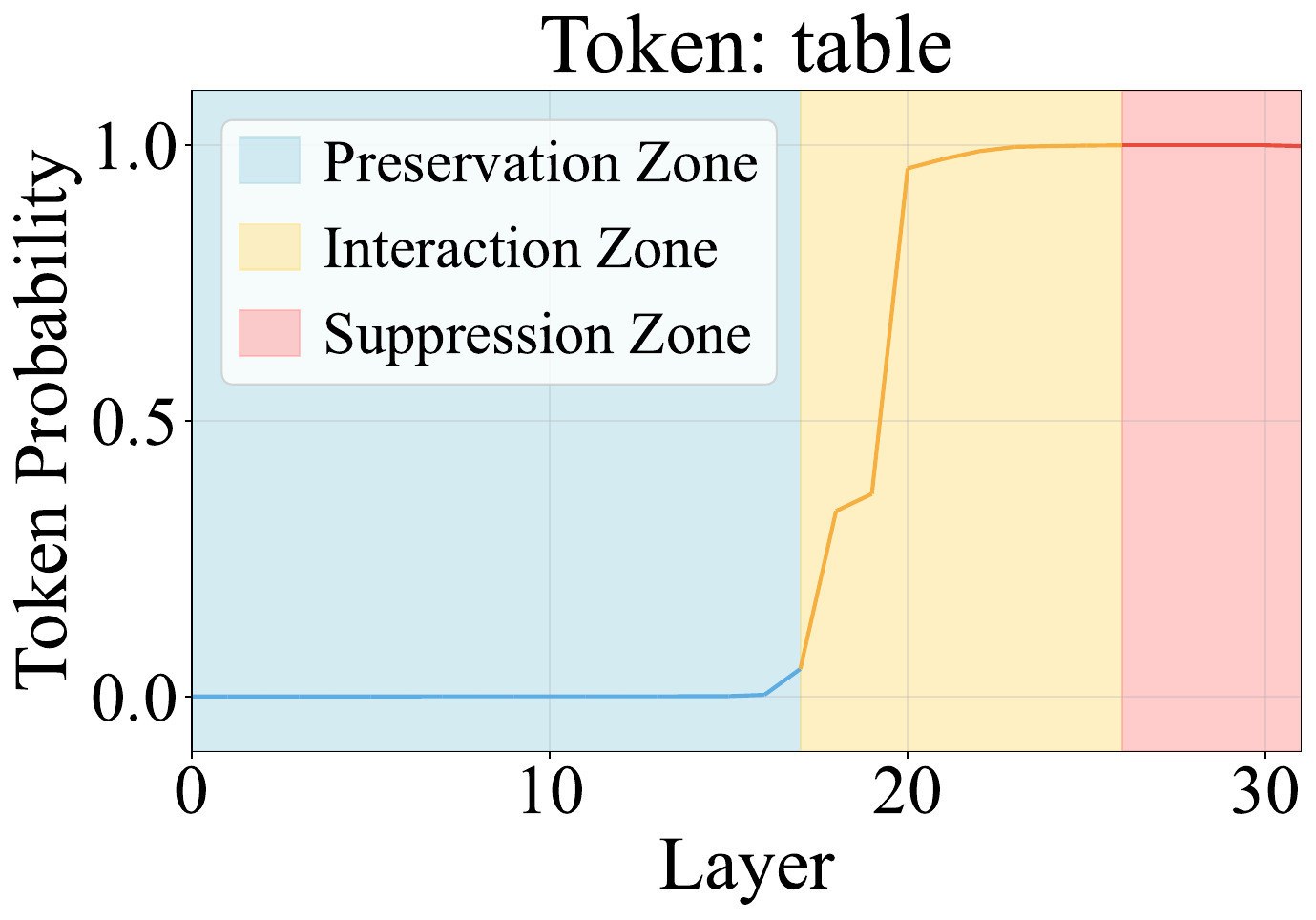}
    \subcaption{(a\textsubscript{3})}
  \end{subfigure}
  \begin{subfigure}{0.3\textwidth}
    \centering
    \includegraphics[width=\linewidth]{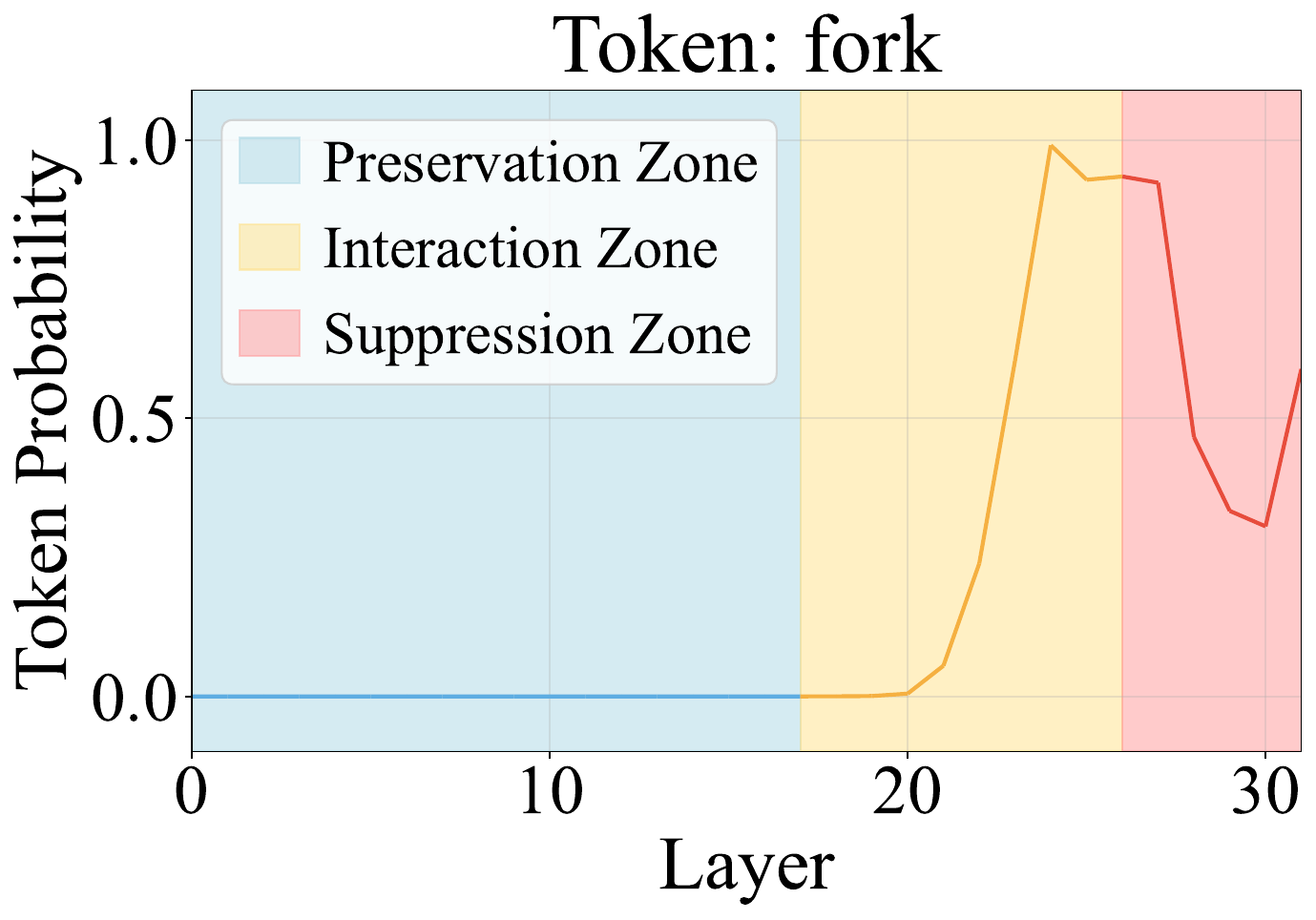}
    \subcaption{(b\textsubscript{1})}
  \end{subfigure}
  \hfill
  \begin{subfigure}{0.3\textwidth}
    \centering
    \includegraphics[width=\linewidth]{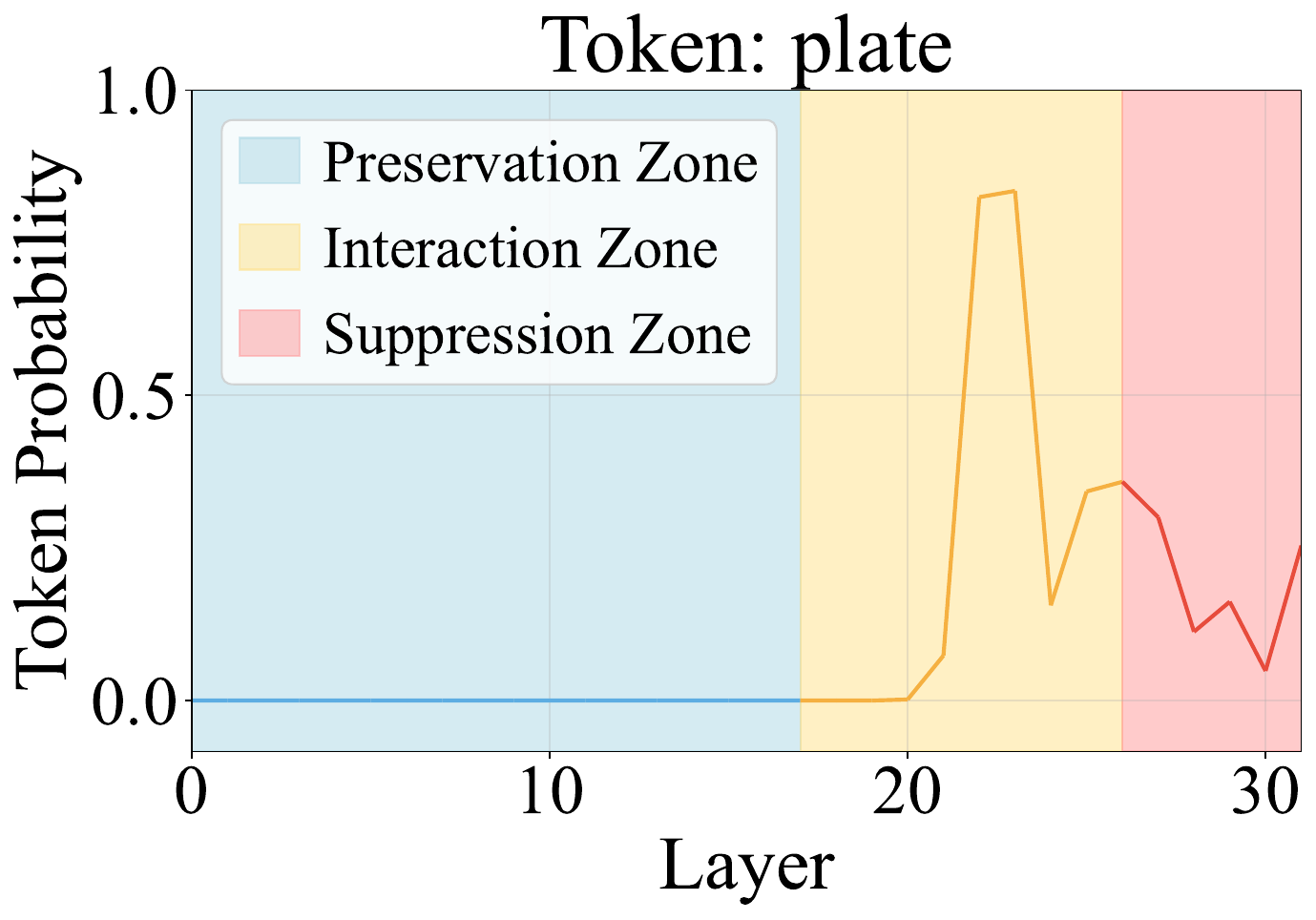}
    \subcaption{(b\textsubscript{2})}
  \end{subfigure}
  \hfill
  \begin{subfigure}{0.3\textwidth}
    \centering
    \includegraphics[width=\linewidth]{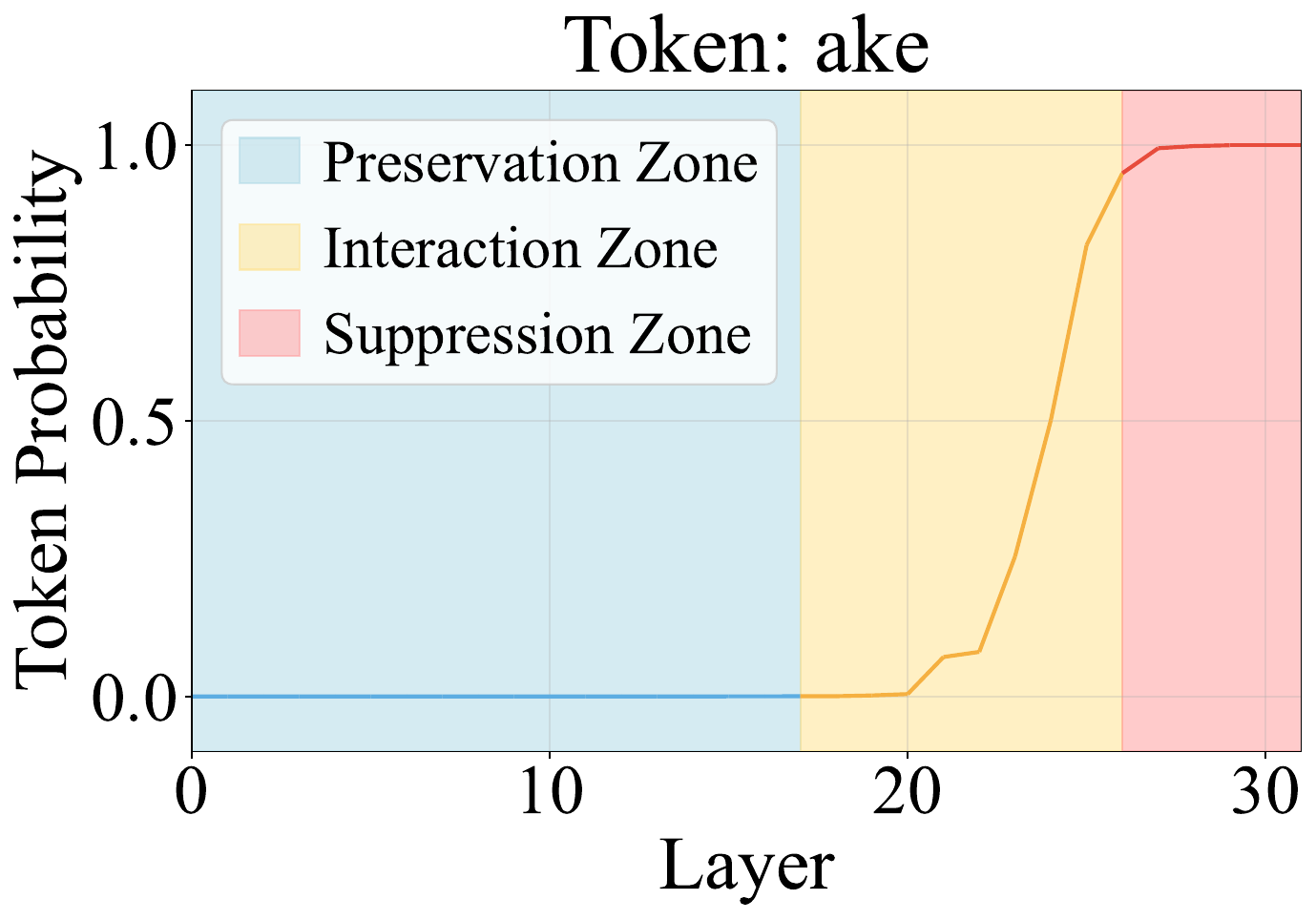}
    \subcaption{(b\textsubscript{3})}
  \end{subfigure}
  \caption{\textbf{Layer-wise token probabilities.} The token
probability is calculated by applying the Softmax function to the
raw logits of the token at each layer. Subfigures
(a\textsubscript{1})--(a\textsubscript{3}) show hallucinated tokens;
subfigures (b\textsubscript{1})--(b\textsubscript{3}) show
non-hallucinated tokens. Y-axis: token probability; X-axis: layer
index (0--31).}
  \label{fig:layer-token-probabilities}
\end{figure*}

\section{Methodology}

\subsection{Why Layer Matters: A Hierarchical Perspective}

MLLMs produce textual responses by progressively transforming
internal representations through a stack of transformer layers.
While prior work \cite{DBLP:conf/cvpr/0009YHS25} has primarily
examined where modality fusion occurs, we shift focus to how
grounded and hallucinated contents propagate through different
layers during decoding.

\begin{figure}[t]
    \centering
    \includegraphics[width=0.9\linewidth]{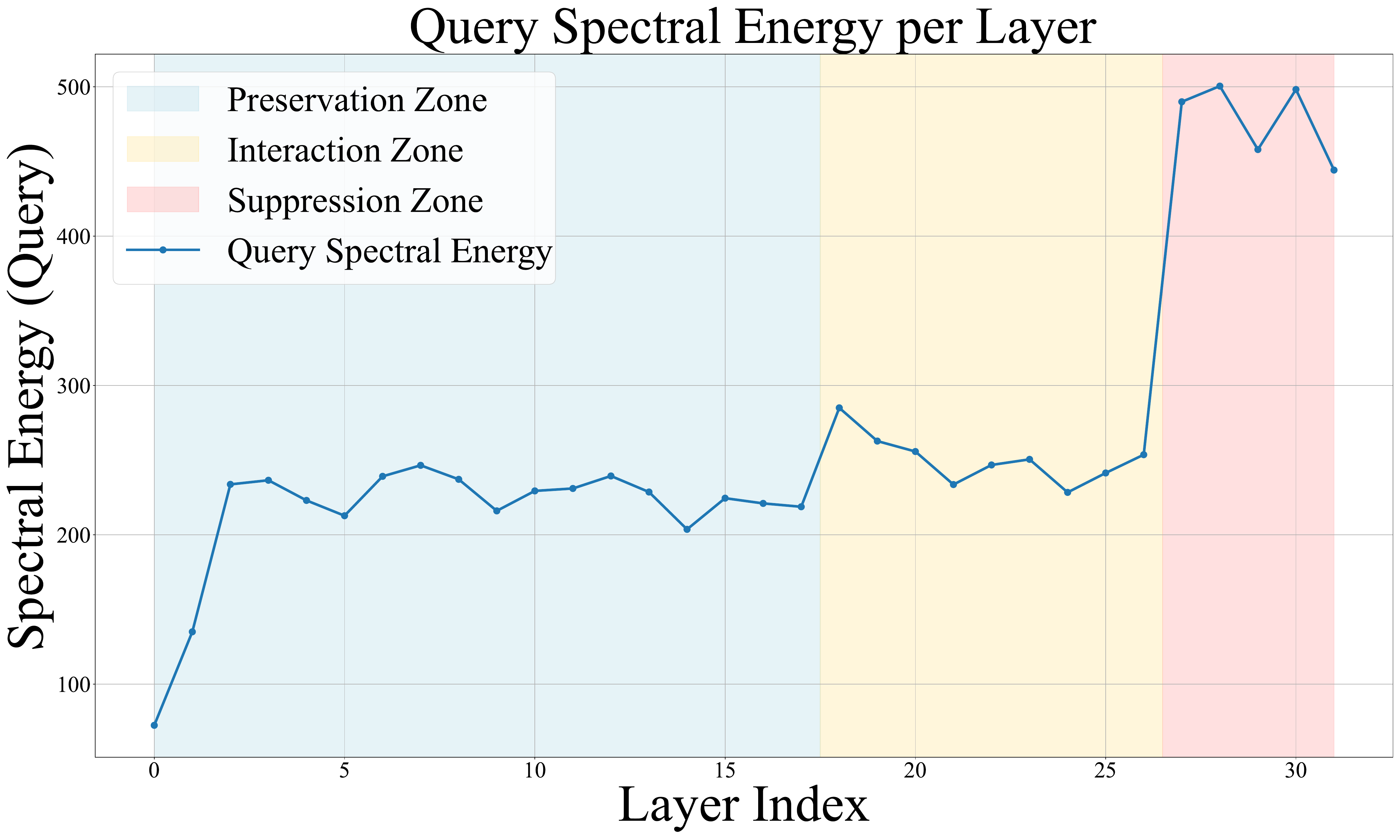}
    \caption{\textbf{Layer-wise spectral energy during token
prediction.} A representative example from a Multimodal Large
Language Model (MLLM) shows that query spectral energy varies across
layers for a single token generation step, forming three zones:
\textit{Preservation} (blue) retains input signals;
\textit{Interaction} (yellow) builds semantic fusion;
\textit{Suppression} (red) shows spikes linked to hallucination.
This pattern motivates layer-wise decoding strategies.}
    \label{fig:spectral-layer-curve}
\end{figure}

\begin{figure}[t]
    \centering
    \includegraphics[width=\linewidth]{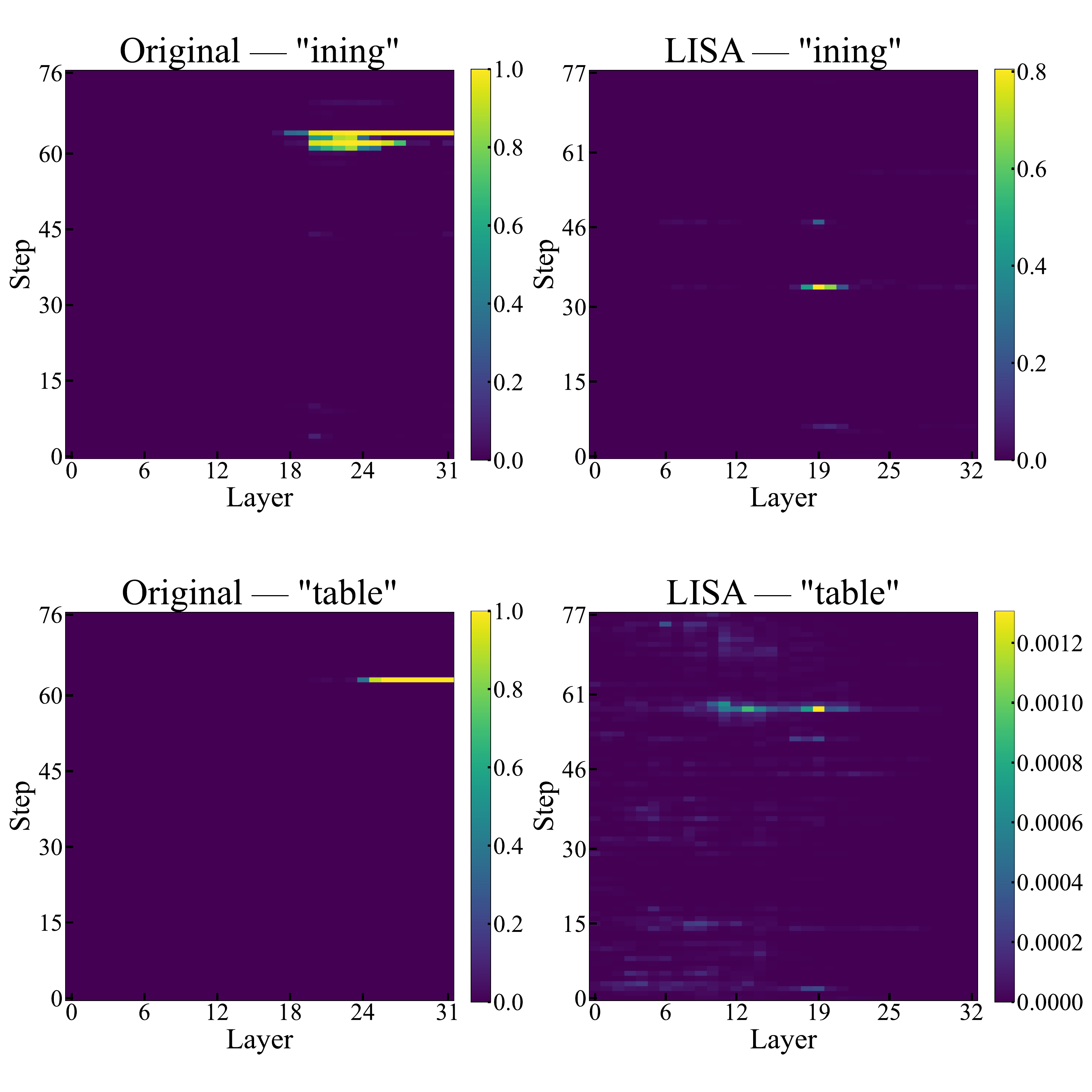}
    \caption{\textbf{Layer-wise heatmaps of hallucinated tokens.}
\textit{Left:} Greedy decoding shows sharp final-layer spikes (e.g.,
``ining'', ``table''). \textit{Right:} LISA suppresses unstable
activations and distributes confidence across layers.}
    \label{fig:token_trajectories}
\end{figure}

\begin{figure*}[t]
    \centering
    \includegraphics[width=\textwidth]{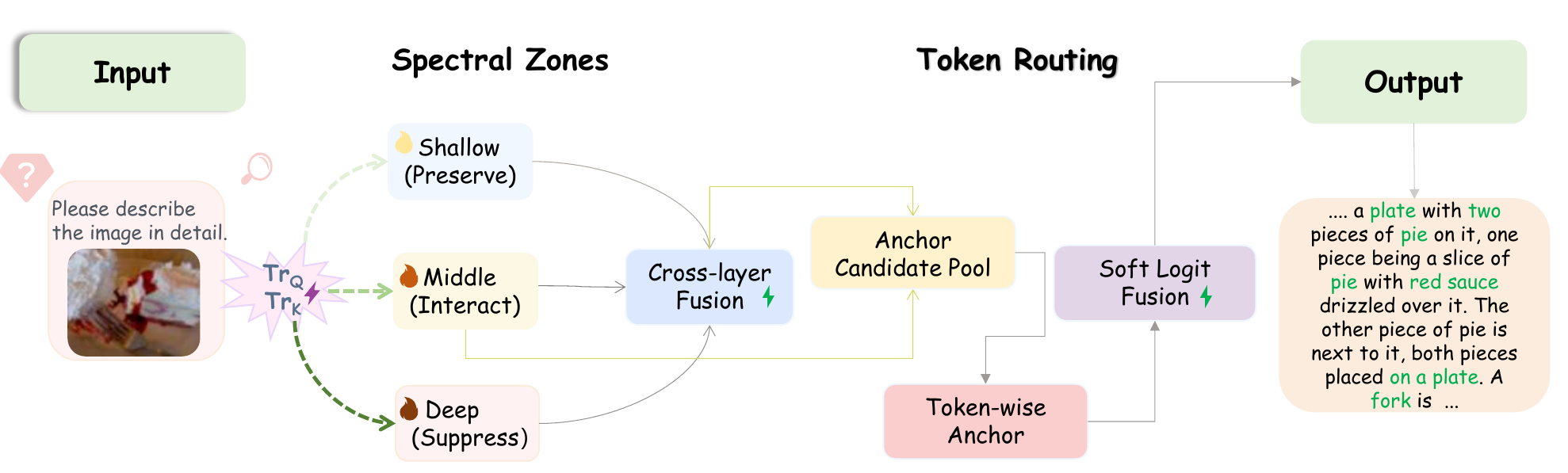}
\caption{\textbf{Overview of LISA.} LISA stabilizes multimodal
generation by modulating the layer-wise spectral energy of
transformer attention. It partitions layers into three spectral
zones---preservation, interaction, and suppression---reflecting the
layer-wise spectral energy progression of queries and keys.
Layer-wise spectral suppression dynamically scales attention to
dampen unstable deep-layer spikes while preserving shallow and
middle-layer semantics. Cross-layer token fusion aggregates stable
representations across selected anchor layers, weighted by spectral
stability. Finally, token-wise anchor selection and soft logit
fusion adaptively integrate multi-layer signals during decoding,
ensuring each token draws from the most stable layers. Together,
LISA combines spectral modulation, anchor-based fusion, and
token-wise routing to mitigate hallucinations while retaining
layer-wise information.}
    \label{fig:layerwise}
\end{figure*}

Figure~\ref{fig:layer-token-probabilities} illustrates
representative layer-wise token probabilities for hallucinated and
non-hallucinated examples. Consistent with prior findings
\cite{DBLP:conf/iclr/Wang00TXDC25}, hallucinated tokens often
maintain persistently high probabilities in deeper layers,
amplifying spurious signals. Interestingly, valid words that are
split into multiple sub-tokens (e.g., cake $\rightarrow$ ``c'',
``ake'') can also show similar high-probability plateaus, even when
they are correctly grounded. This phenomenon suggests that
persistent activations in deeper layers do not always imply
hallucinations, underscoring the importance of a robust, layer-wise
mechanism that can dampen unstable signals without over-suppressing
legitimate content.

To interpret this behavior further, we plot a representative query
spectral energy curve in Figure~\ref{fig:spectral-layer-curve}. We
observe that in the shallow layers, the spectral energy remains low
and dispersed. This zone primarily serves to preserve the input
features, acting as a robust foundation for subsequent processing.
As signals propagate through the middle layers, a steady increase in
spectral energy indicates the progressive integration of semantic
information. This is where the model performs crucial cross-modal
fusion, blending visual and textual cues to build a coherent,
grounded representation of the input. However, a distinct and
problematic pattern emerges in the deeper layers. Here, we
frequently observe abrupt spikes in spectral energy that correlate
with the model's confidence in specific tokens. Consistent with
prior work on large models such as LLaVA
\cite{DBLP:conf/cvpr/LiuLLL24}, these spikes often align with the
model's overconfident predictions, leading to the generation of
spurious or hallucinatory content. While some of these spikes may
correspond to legitimate phenomena like token splits for valid
words, they more typically signal the amplification of unstable
representations. This observation underscores a critical challenge:
distinguishing between beneficial amplification of grounded signals
and detrimental amplification of spurious ones. This is why LISA
adopts a layer-wise spectral modulation strategy. Rather than
rigidly distinguishing each case, it adaptively controls layer-wise
signals to mitigate these deep-layer activations while preserving
valid structures from more stable shallow and middle layers.

Figure~\ref{fig:token_trajectories} demonstrates the practical
impact of our spectral modulation. Compared with existing MLLMs
\cite{DBLP:conf/cvpr/LiuLLL24,DBLP:conf/nips/Dai0LTZW0FH23,DBLP:journals/corr/abs-2308-12966,DBLP:journals/corr/abs-2502-13923},
LISA effectively suppresses final-layer overactivation for
hallucinated tokens and promotes a more balanced layer-wise
activation pattern, resulting in more reliable and faithful outputs.

Together, these observations suggest a functional hierarchy: shallow
layers preserve inputs; middle layers integrate grounded semantics;
deep layers tend to amplify spurious signals. Motivated by this
hierarchical view, we propose LISA, which explicitly partitions the
transformer into three spectral zones---preservation, interaction,
and suppression, as shown in Figure~\ref{fig:layerwise}. By
combining layer-wise spectral modulation, cross-layer fusion, and
token-wise anchor routing, LISA adaptively balances final-layer
representation capacity with stability, mitigating hallucinations
without retraining or changing parameters of MLLMs.

\subsection{Spectral Modulation for Layer-wise Stability}

Motivated by prior insights \cite{DBLP:conf/iclr/TangHLSYL25} that
controlling the query-key eigenspectrum can stabilize generation, we
introduce LISA's first core mechanism: \textit{Spectral Modulation
(SM)}. It combines layer-wise spectral suppression and cross-layer
token fusion to mitigate hallucinations in multimodal generation.
This approach builds on the observation that transformer layers
exhibit layer-wise functional roles, and that different layers
contribute differently to the final output.

\paragraph{Layer-wise Spectral Suppression.}

For each transformer layer $l$, we define its spectral energy as the
squared Frobenius norm of its query and key matrices:
\begin{equation}
\begin{array}{l}
\mathrm{Tr}_Q^{(l)} = \|\mathbf{Q}^{(l)}\|_F^2 = \mathrm{Tr}(\mathbf{Q}^{(l)} \mathbf{Q}^{(l)\top}),\\
\mathrm{Tr}_K^{(l)} = \|\mathbf{K}^{(l)}\|_F^2 = \mathrm{Tr}(\mathbf{K}^{(l)} \mathbf{K}^{(l)\top}).
\end{array}
\label{eq:spectral_energy}
\end{equation}
As shown in Figure~\ref{fig:spectral-layer-curve}, spectral energy
remains low and dispersed in shallow layers (preservation zone),
gradually increases in middle layers (interaction zone) as semantic
fusion occurs, and often shows abrupt spikes in deep layers
(suppression zone), which we have found to align with the emergence
of hallucinations.

To suppress unstable activations and prevent them from dominating
the generation process, we apply a dynamic spectral modulation
factor to the scaled dot-product attention. This mechanism is
designed to constrain high-energy spikes while minimally affecting
stable, low-energy activations:
\begin{equation}
\begin{array}{l}
\lambda_Q^{(l)} = 1 + \frac{\gamma^{(l)}}{\log(\mathrm{Tr}_Q^{(l)} + \epsilon)},\\
\lambda_K^{(l)} = 1 + \frac{\gamma^{(l)}}{\log(\mathrm{Tr}_K^{(l)} + \epsilon)},
\end{array}
\label{eq:spectral_suppression}
\end{equation}
where $\epsilon$ is a small positive constant to ensure numerical
stability. The suppression strength parameter, $\gamma$, is not a
fixed value but is dynamically set based on the layer index $l$. We
model this relationship using a simple monotonically increasing
function that reflects the functional hierarchy of the transformer.
\begin{equation}
\gamma^{(l)} = \gamma_{\text{min}} + (\gamma_{\text{max}} -
\gamma_{\text{min}}) \cdot \frac{l}{L},
\label{eq:gamma}
\end{equation}
where $L$ is the total number of layers in the model. For the model
used in our experiments, $L$ is the total number of decoder layers,
which is 32. This linear model is the simplest and most
interpretable way to capture the observed monotonic increase in
representation change across layers. It requires tuning only two
boundary parameters, $\gamma_{\text{min}}$ and
$\gamma_{\text{max}}$. Specifically, setting $\gamma_{\text{min}} =
0.0$ ensures the shallowest layers apply no spectral suppression at
all, preserving input fidelity. $\gamma_{\text{max}} = 0.9$ is
empirically chosen as the maximum strength that significantly
curtails hallucination while maintaining coherent generation
quality. This dynamic approach ensures that the suppression strength
increases progressively with layer depth, ranging from a minimum
value $\gamma_{\text{min}}$ for the shallowest layers to a maximum
value $\gamma_{\text{max}}$ for the deepest layers. The modulated
attention then becomes:
\begin{equation}
\mathbf{A}^{(l)} = \text{Softmax}\left(\frac{(\lambda_Q^{(l)}
\mathbf{Q}^{(l)}) (\lambda_K^{(l)}
\mathbf{K}^{(l)})^\top}{\sqrt{d_k}}\right),
\label{eq:spectral_attention}
\end{equation}
where $d_k$ is the dimensionality of the query/key vectors. By
specifically constraining high-energy spikes in deeper layers, this
targeted suppression method mitigates overactivation while
preserving the robust representation capacity of the network's more
stable layers.

\paragraph{Cross-layer Token Fusion.}

Beyond local suppression, LISA globally aggregates stable
information through its second core mechanism, \textit{Cross-layer
Fusion (CF)}. Rather than relying solely on the final layer or
averaging across all layers, CF strategically fuses token states
from a carefully selected anchor layer set $\mathcal{A}$, which
consists of representative layers sampled from the three zones. To
guide this fusion, we introduce \textit{spectral stability} to
measure the reliability of layer representations. It is defined as
the inverse of the spectral energy:
\begin{equation}
s^{(l)} = \frac{1}{\mathrm{Tr}_Q^{(l)} + \mathrm{Tr}_K^{(l)} +
\epsilon}, \label{eq:spectral_stability}
\end{equation}
where $Tr_Q^{(l)}$ and $Tr_K^{(l)}$ denote the spectral trace
energies of the query and key matrices, respectively. This metric
captures the layer-wise reliability: low energy indicates stable,
grounded representations, while high energy indicates volatile,
potentially hallucinated representations. Our fusion mechanism
leverages this by amplifying stable signals from shallow/middle
layers and dampening spurious deep-layer activations. We acknowledge
that this strategy may weakly regularize some high-confidence
legitimate sub-tokens, but empirically, the resulting net reduction
in hallucination significantly outweighs this minimal trade-off.

The fusion weights are normalized to ensure a consistent
contribution:
\begin{equation}
\alpha^{(l)} = \frac{s^{(l)}}{\sum_{j \in \mathcal{A}} s^{(j)}},
\quad \sum_{l \in \mathcal{A}} \alpha^{(l)} = 1.
\label{eq:layerwise_fusion_weights}
\end{equation}
The final fused representation, $\mathbf{H}_{\text{LISA}}$, is
formulated as a weighted sum of the representations from the anchor
layers:
\begin{equation}
\mathbf{H}_{\text{LISA}}
= \sum_{l \in \mathcal{A}} \alpha^{(l)}\, \mathbf{H}^{(l)}.
\label{eq:layerwise_fusion}
\end{equation}
This strategic aggregation amplifies grounded signals from stable
layers while dampening unstable ones, leading to more reliable
generation outputs.

\subsection{Layer-wise Token Integration for Stable Generation}

Building on stable multi-layer signals, LISA extends this layer-wise
design into the decoding stage through its third core mechanism,
which we term \textit{Token-wise Soft Fusion (TSF)}. Unlike
conventional pipelines that rely solely on final-layer logits, TSF
aligns the generation of each token with the layer-wise spectral
pattern: grounded content stabilizes in shallow and intermediate
layers, while hallucinations often correlate with unstable
deep-layer spikes.

\paragraph{Token-wise Anchor Selection.}

To leverage this, we first partition the transformer into three
zones---preservation, interaction, and suppression---and select
representative layers from each zone to form an initial anchor set
$\mathcal{A}_{\text{anchor}}$. Crucially, we retrieve the hidden
states from these intermediate layers and pass them to the model's
shared final classification head ($\text{LM Head}$) to obtain
layer-wise logits ($\mathbf{L}_{\mathbf{H}, t}$) and token
probabilities. The cross-layer fused representation
\(\mathbf{H}_{\text{LISA}}\) is computed from this set (as defined
in Eq.~\ref{eq:layerwise_fusion}), and is then included as an
additional candidate for the final token-wise fusion.

The complete set of candidate anchors for logit fusion is therefore:
\begin{equation}
\mathcal{C} = \{\mathbf{H}^{(l)} \mid l \in \mathcal{A}_{\text{anchor}}\} \cup
\{\mathbf{H}_{\text{LISA}}\}.
\label{eq:candidate_set}
\end{equation}
Since $\mathbf{H}_{\text{LISA}}$ is a fused representation and does
not correspond to a single layer index, we define its effective
stability, $s(\mathbf{H}_{\text{LISA}})$, as the average spectral
stability of its constituent anchor layers:
\begin{equation}
s(\mathbf{H}_{\text{LISA}}) =
\frac{1}{|\mathcal{A}_{\text{anchor}}|} \sum_{l \in
\mathcal{A}_{\text{anchor}}} s^{(l)}.
\end{equation}
For \textbf{each token $c$ considered for generation}, the optimal
anchor representation $\mathbf{H}^*_c$ is dynamically selected by
maximizing spectral stability-weighted probability:
\begin{equation}
\mathbf{H}^*_c = \arg\max_{\mathbf{H} \in \mathcal{C}} \left(
s(\mathbf{H}) \cdot \text{Softmax}(\mathbf{L}_{\mathbf{H}, t})_c
\right),
\label{eq:optimal_anchor_selection}
\end{equation}
where $\mathbf{L}_{\mathbf{H}, t}$ is the logit vector obtained by
passing $\mathbf{H}$ through the $\text{LM Head}$. The term
$\mathbf{H}^*_c$ is the optimal anchor selected specifically for
token $c$, and $s(\mathbf{H})$ is the spectral stability of the
source of $\mathbf{H}$ (either $s^{(l)}$ for a single layer or
$s(\mathbf{H}_{\text{LISA}})$ for the fused representation). This
process ensures that the chosen anchor $\mathbf{H}^*_c$ is the
representation that is most stable and provides the highest
confidence for the intended token $c$, aligning the calculation with
the mechanism's core goal.

\paragraph{Soft Logit Fusion.}

Finally, token logits are fused to balance representation capacity
and stability. Since the optimal anchor is token-dependent, the
fusion is performed for every candidate token $c$:
\begin{equation}
\hat{\mathbf{L}}_t(c) =  \mathbf{L}_{L,t}(c) +\beta\, \mathbf{L}_{\mathbf{H}^*_c, t}(c), \quad \beta \in [0, 1].
\label{eq:soft_logit_fusion}
\end{equation}
Here, $\hat{\mathbf{L}}_t(c)$ is the resulting fused logit for token
$c$. The terms $\mathbf{L}_{L,t}(c)$ and
$\mathbf{L}_{\mathbf{H}^*_c, t}(c)$ are the logits for token $c$
from the final layer ($L$) and its selected token-specific anchor
($\mathbf{H}^*_c$), respectively. $\beta$ controls the trade-off
between the capacity of the final layer and the stability of the
selected anchor. This design allows each token to adaptively draw
from its own most stable, high-confidence source, thereby mitigating
hallucinations across decoding strategies while preserving
layer-wise information.

LISA integrates multi-layer signals, enhancing the stability of
token generation. By dynamically selecting the most stable anchor
layer based on both spectral stability and token probability, and
soft-fusing the logits, LISA minimizes hallucinations and ensures
the generation of meaningful, coherent content. Together, these
mechanisms yield a robust decoding approach that enhances both
accuracy and stability, while preserving layer-wise information
throughout the generation process.

\section{Experiments}

\subsection{Experimental Settings}

\noindent\textbf{Datasets.} We use MSCOCO
\cite{DBLP:conf/eccv/LinMBHPRDZ14}, which contains diverse everyday
scenes with human-annotated captions. Only images are used for
inference; captions are for evaluation.

\noindent\textbf{Baselines.} We compare LISA with standard decoding
strategies (e.g., greedy decoding, beam search, and nucleus
sampling) and recent decoding-time hallucination mitigation methods,
including DoLa \cite{DBLP:conf/iclr/ChuangXLKGH24}, DeCo
\cite{DBLP:conf/iclr/Wang00TXDC25}, VCD
\cite{DBLP:conf/cvpr/LengZCLLMB24}, and OPERA
\cite{DBLP:conf/cvpr/HuangDZ0H0L0Y24}.

\noindent\textbf{Backbone Models.} We evaluate our method with four
MLLMs: LLaVA \cite{DBLP:conf/cvpr/LiuLLL24}, InstructBLIP
\cite{DBLP:conf/nips/Dai0LTZW0FH23}, Qwen-VL
\cite{DBLP:journals/corr/abs-2308-12966}, and Qwen2.5-VL
\cite{DBLP:journals/corr/abs-2502-13923}, all used in frozen form
without fine-tuning.

\noindent\textbf{Implementation Details.} For a 32-layer
decoder-only language model, the dynamic spectral suppression
parameter $\gamma^{(l)}$ is governed by
$\mathbf{\gamma_{\text{min}}=0.0}$ and
$\mathbf{\gamma_{\text{max}}=0.9}$ (validated in
Figure~\ref{fig:gamma}).

To ensure reproducibility, we formally define the three functional
zones (defined by spectral energy) using a proportional split based
on the total layer count $L$:
\begin{itemize}
    \item \textbf{Preservation Zone:} $0 \le l \le \lfloor 0.30 \cdot L
    \rfloor$;
    \item \textbf{Interaction Zone:} $\lfloor 0.30 \cdot L \rfloor < l \le \lfloor 0.85 \cdot L
    \rfloor$;
    \item \textbf{Suppression Zone:} $\lfloor 0.85 \cdot L \rfloor < l \le
    L$.
\end{itemize}
This layer-wise hierarchical division is consistent with findings
that deeper layers are more susceptible to generating non-factual
content \cite{DBLP:conf/iclr/Wang00TXDC25}. Main decoding
hyperparameters are given in Table~\ref{tab:decode-hparams}.

\begin{table}[t] \footnotesize
\centering \resizebox{1.0\columnwidth}{!}{
\begin{tabular}{|l|c|l|}
\hline
\textbf{Hyperparameter} & \textbf{Value} & \textbf{Description} \\
\hline
Beam size             & 5         & Beam size \\
Temperature           & 0.7       & Sampling temperature \\
Max tokens            & 512       & Maximum generation length \\
$\beta$               & 0.6       & Token-wise fusion weight \\
$\epsilon$            & $10^{-7}$ & Spectral stabilizer \\
$\gamma_{\text{min}}$ & 0.0       & Minimum suppression strength for $\gamma$ \\
$\gamma_{\text{max}}$ & 0.9       & Maximum suppression strength for $\gamma$ \\
\hline
\end{tabular}}
\caption{Main hyperparameters used in LISA. The value of $\gamma$ is
dynamically set based on the function defined in
Eq.~\ref{eq:gamma}.} \label{tab:decode-hparams}
\end{table}

\begin{table*}[t] \footnotesize
\centering
\begin{tabular}{llcccccc}
\hline \textbf{Decoding} & \textbf{Method} &
\multicolumn{2}{c}{\textbf{LLaVA-1.5}} &
\multicolumn{2}{c}{\textbf{InstructBLIP}}
& \multicolumn{2}{c}{\textbf{Qwen-VL}} \\
& & CHAIR$_\text{S}$$\downarrow$ & CHAIR$_\text{I}$$\downarrow$ & CHAIR$_\text{S}$$\downarrow$ & CHAIR$_\text{I}$$\downarrow$ & CHAIR$_\text{S}$$\downarrow$ & CHAIR$_\text{I}$$\downarrow$\\
\hline
Greedy   & Vanilla       & 45.0 & 14.7 & 58.8 & 23.7 & 46.0 & 12.5 \\
         & DoLa          & 47.8 & 13.8 & 48.4 & 15.9 & 46.8 & 12.9 \\
         & DeCo          & 37.8 & 11.1 & 41.2 & 14.4 & 42.2 & 10.7 \\
         & \textbf{LISA} & \textbf{34.2} {\scriptsize $\downarrow$10.8} & \textbf{10.4} {\scriptsize $\downarrow$4.3} & \textbf{31.6} {\scriptsize $\downarrow$27.2} & \textbf{10.9} {\scriptsize $\downarrow$12.8} & \textbf{29.0} {\scriptsize $\downarrow$17.0} & \textbf{9.2} {\scriptsize $\downarrow$3.3} \\
\hline
Beam     & Vanilla       & 48.8 & 13.9 & 55.6 & 15.8 & 41.8 & 10.8 \\
         & OPERA         & 44.6 & 12.8 & 46.4 & 14.2 & 34.6 & 9.5 \\
         & DeCo          & 33.0 & 9.7  & 43.8 & 12.7 & 32.0 & 8.7 \\
         & \textbf{LISA} & \textbf{29.2} {\scriptsize $\downarrow$19.6} & \textbf{9.1} {\scriptsize $\downarrow$4.8} & \textbf{31.0} {\scriptsize $\downarrow$24.6} & \textbf{11.6} {\scriptsize $\downarrow$1.1} & \textbf{20.4} {\scriptsize $\downarrow$21.4} & \textbf{7.4} {\scriptsize $\downarrow$3.4} \\
\hline
Nucleus  & Vanilla       & 48.8 & 14.2 & 54.6 & 24.8 & 49.2 & 13.1 \\
         & VCD           & 54.0 & 16.0 & 58.0 & 17.0 & 46.4 & 11.9 \\
         & DeCo          & 42.8 & 13.2 & 43.6 & \textbf{12.9} {\scriptsize $\downarrow$11.9} & 43.8 & 11.8 \\
         & \textbf{LISA} & \textbf{39.0} {\scriptsize $\downarrow$9.8} & \textbf{11.6} {\scriptsize $\downarrow$2.6} & \textbf{39.4} {\scriptsize $\downarrow$15.2} & {16.3} & \textbf{30.2} {\scriptsize $\downarrow$19.0} & \textbf{10.3} {\scriptsize $\downarrow$2.8} \\
\hline
\end{tabular}
\caption{\textbf{CHAIR hallucination evaluation results} for
\textbf{LLaVA-1.5}, \textbf{InstructBLIP}, and \textbf{Qwen-VL}
across different decoding strategies. Lower scores on
CHAIR$_\text{S}$ and CHAIR$_\text{I}$ indicate fewer hallucinations.
OPERA is a beam-search-based method; VCD is designed for nucleus
sampling; DeCo is a general decoding-compatible approach. LISA
denotes our method, and is applied consistently across all
settings.} \label{tab:chair-lisa-final}
\end{table*}

\begin{table}[t] \footnotesize
\centering \resizebox{1.0\columnwidth}{!}{
\begin{tabular}{llcccc}
\hline \textbf{Decoding} & \textbf{Method}
& \textbf{CHAIR$\downarrow$} & \textbf{Cover$\uparrow$} & \textbf{Hal.$\downarrow$} & \textbf{Cog.$\downarrow$} \\
\hline
Greedy   & Vanilla       & 8.2 & 48.9 & 34.3 & 4.0 \\
         & DoLa          & 8.0 & 50.8 & 37.5 & 4.3 \\
         & DeCo          & 6.6 & 47.5 & 28.1 & 2.8 \\
         & \textbf{LISA} & \textbf{6.5} {\scriptsize $\downarrow$1.7} & \textbf{47.2} {\scriptsize $\downarrow$1.7} & \textbf{23.1} {\scriptsize $\downarrow$11.2} & \textbf{1.9} {\scriptsize $\downarrow$2.1} \\
\hline
Beam     & Vanilla       & 7.1 & 50.7 & 32.4 & 3.8 \\
         & OPERA         & 6.4 & 49.0 & 27.5 & 2.9 \\
         & DeCo          & 6.3 & 46.8 & 25.1 & 2.4 \\
         & \textbf{LISA} & \textbf{5.9} {\scriptsize $\downarrow$1.2} & \textbf{45.6} {\scriptsize $\downarrow$5.1} & \textbf{21.9} {\scriptsize $\downarrow$10.5} & \textbf{2.0} {\scriptsize $\downarrow$1.8} \\
\hline
Nucleus  & Vanilla       & 10.2 & 50.2 & 43.3 & 4.5 \\
         & VCD           & 9.0  & 51.7 & 40.2 & 4.4 \\
         & DeCo          & 8.3  & 48.0 & 37.5 & 3.4 \\
         & \textbf{LISA} & \textbf{7.2} {\scriptsize $\downarrow$3.0} & \textbf{47.6} {\scriptsize $\downarrow$2.6} & \textbf{28.8} {\scriptsize $\downarrow$14.5} & \textbf{2.6} {\scriptsize $\downarrow$1.9} \\
\hline
\end{tabular}}
\caption{AMBER evaluation results of \textbf{LLaVA-1.5} under
different decoding strategies. LISA consistently reduces
hallucination-related metrics---CHAIR (hallucinated object rate),
Hal. (rate of hallucinated responses), and Cog. (alignment with
human cognitive biases)---while incurring a slight decrease in
Cover, which quantifies the proportion of ground-truth objects
correctly mentioned in the response.} \label{tab:amber-llava}
\end{table}

\subsection{Evaluation Benchmarks}

\textbf{CHAIR.} The Caption Hallucination Assessment with Image
Relevance (CHAIR) \cite{DBLP:conf/emnlp/RohrbachHBDS18} evaluates
object hallucination in captions by comparing generated object
mentions with ground-truth labels. It is defined as:
\begin{align*}
\mathrm{CHAIR}_{\mathrm{I}} &= \frac{\big|\,\text{Hallucinated Objects}\,\big|}{\big|\,\text{Mentioned Objects}\,\big|}, \\
\mathrm{CHAIR}_{\mathrm{S}} &= \frac{\big|\,\text{Captions with Hallucinations}\,\big|}{\big|\,\text{All Captions}\,\big|}.
\end{align*}
Following \citet{DBLP:conf/cvpr/HuangDZ0H0L0Y24}, we evaluate on 500
MSCOCO 2014 validation images, averaging the results over 3 runs
with different random seeds, using the caption prompt:
\textit{``Please describe the image in detail.''}

\textbf{AMBER.} AMBER \cite{DBLP:journals/corr/abs-2311-07397}
evaluates hallucinations in image captions by assessing hallucinated
object rate (CHAIR), coverage of ground-truth objects (Cover),
proportion of hallucinated responses (Hal.), and tendency to mention
cognitively biased objects (Cog.). Lower scores for CHAIR, Hal., and
Cog. indicate fewer hallucinations, while a higher Cover score
reflects stronger grounding.

\textbf{MME.} The MME benchmark
\cite{DBLP:journals/corr/abs-2306-13394} evaluates the general
capabilities of multimodal large language models across 14
subskills, including OCR, object recognition, spatial relations, and
commonsense reasoning. It uses multiple-choice questions on natural
images. We follow the official protocol and report overall accuracy.

\textbf{POPE.} The Polling-based Object Probing Evaluation (POPE)
\cite{DBLP:conf/emnlp/LiDZWZW23} assesses hallucination through a
binary question-answering format. For each image, the model answers
whether a queried object exists in the scene. Performance is
measured using the standard F1 score.

To evaluate robustness, POPE splits queried objects into three
subsets: random (arbitrary samples), popular (frequent objects), and
adversarial (visually similar to ground-truth). Each subset includes
500 MSCOCO images with six binary questions per image.

\subsection{Performance Analysis}

We evaluate LISA using the four benchmarks introduced above---CHAIR,
AMBER, POPE, and MME---which together capture diverse facets of
generation reliability, including hallucination mitigation, visual
grounding, and factual consistency across different decoding
strategies and model families.

\paragraph{Robust Hallucination Mitigation.}

Table~\ref{tab:chair-lisa-final} shows that LISA consistently
reduces hallucinated object mentions (CHAIR$_\text{S}$,
CHAIR$_\text{I}$) across different models and decoding strategies on
the CHAIR benchmark. These improvements are particularly evident
under beam search, where LISA surpasses existing decoding-time
approaches such as DoLa and DeCo, highlighting its effectiveness
under conditions most prone to hallucination errors.

Similarly, as shown in Table~\ref{tab:amber-llava}, LISA
significantly reduces hallucination metrics (Hal. and Cog.) while
preserving grounded object coverage (Cover), demonstrating
robustness even in tasks requiring commonsense and spatial
reasoning.

\begin{table}[t] \footnotesize
\centering
\begin{tabular}{llccc}
\hline \textbf{Method} & \textbf{LLaVA-1.5} & \textbf{InstructBLIP}
& \textbf{Qwen-VL} \\
\hline
Greedy       & 82.2 & 80.0 & 85.2 \\
Nucleus      & 83.1 & 79.8 & 84.5 \\
Beam         & 84.9 & 84.4 & 85.3 \\
DeCo         & 85.4 & 81.8 & 85.2 \\
DoLa         & 83.2 & 83.4 & 85.8 \\
VCD          & 83.1 & 79.9 & 84.7 \\
OPERA        & 85.4 & 84.8 & 86.1 \\
\textbf{LISA} & \textbf{86.7} {\scriptsize $\uparrow$4.5}
             & \textbf{84.9} {\scriptsize $\uparrow$5.1}
             & \textbf{86.7} {\scriptsize $\uparrow$2.2} \\
\hline
\end{tabular}
\caption{POPE F1 scores of \textbf{LLaVA-1.5},
\textbf{InstructBLIP}, and \textbf{Qwen-VL} under various decoding
strategies. Higher scores indicate stronger alignment with object
existence. For fair comparison, LISA results are obtained under
nucleus sampling.} \label{tab:pope-f1}
\end{table}

\paragraph{Improved Visual Grounding and Recall.}

As shown in Table~\ref{tab:pope-f1}, LISA consistently improves F1
scores on the POPE benchmark across all evaluated MLLMs,
demonstrating enhanced grounding to actual object presence. While
initially developed to stabilize caption generation via layer-wise
spectral control, LISA also enhances binary yes/no existence
prediction, demonstrating its generalizability across output
formats. Additional results on Qwen2.5-VL confirm that the
layer-wise design of LISA reliably improves object recall and
grounding under various decoding strategies.

\begin{figure}[t]
    \centering
    \includegraphics[width=0.7\linewidth]{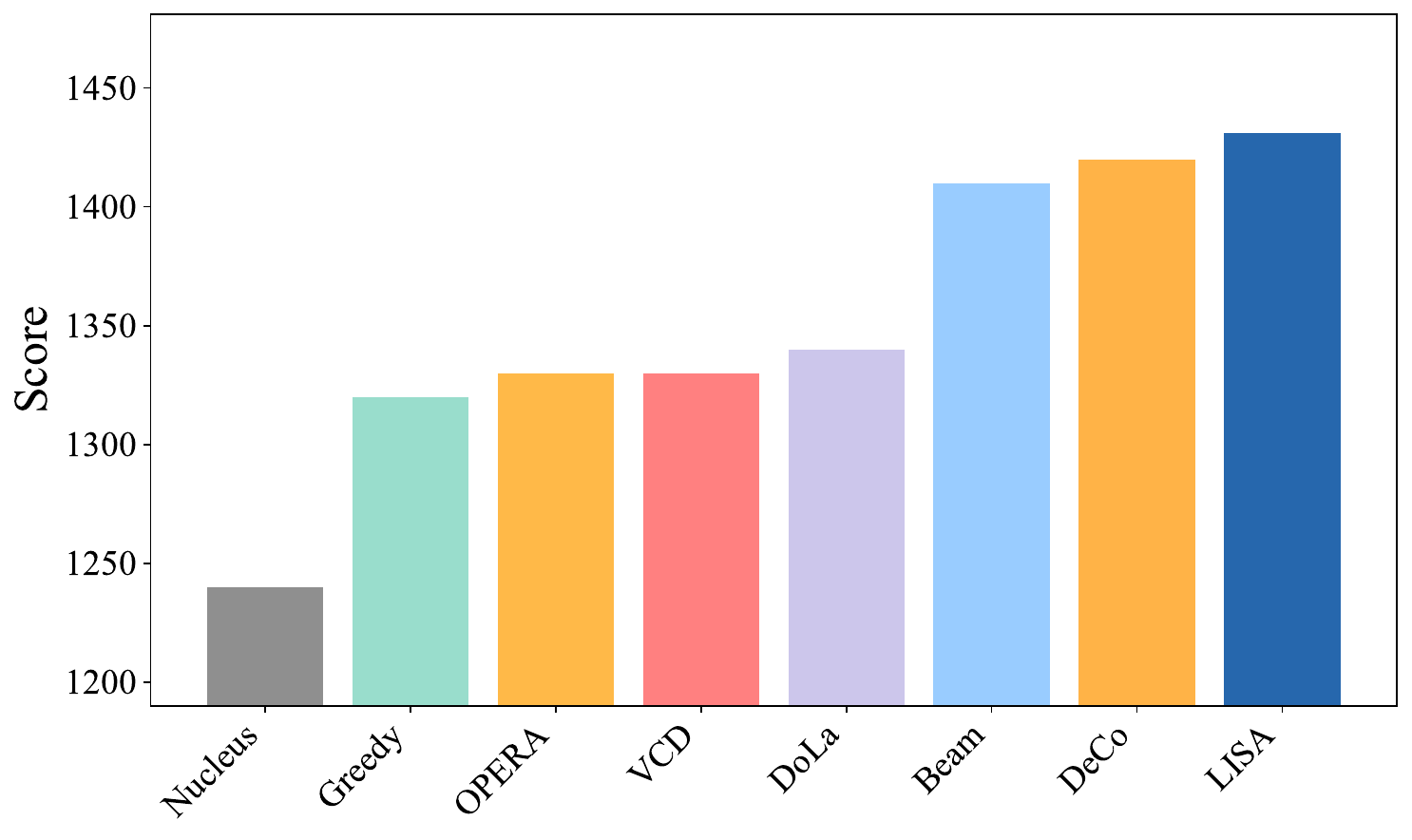}
    \caption{\textbf{MME benchmark results.} LISA outperforms standard decoding
(i.e., greedy, beam, nucleus) and recent mitigation methods (i.e.,
DoLa, VCD, OPERA, DeCo).}
    \label{fig:mme_results}
\end{figure}

\paragraph{Consistent Improvements across Tasks.}

Finally, as shown in Figure~\ref{fig:mme_results}, LISA achieves the
highest overall score on the MME benchmark among recent
decoding-time hallucination mitigation techniques, demonstrating
robust performance for tasks that demand broader multimodal
understanding, including spatial reasoning, attribute alignment, and
fine-grained object grounding.

Across benchmarks and decoding strategies, LISA consistently
enhances visual grounding and generation stability across models,
offering a practical, training-free solution for mitigating
hallucinations in real-world MLLM applications.

\subsection{Hyperparameter Study}

To optimize the LISA framework, we study two critical parameters:
the maximum suppression strength ($\gamma_{\text{max}}$) for
Spectral Modulation (SM) and the soft logit fusion weight ($\beta$)
for Token-wise Soft Fusion (TSF). All experiments here used
LLaVA-1.5.

\paragraph{Effect of Maximum Suppression Strength ($\gamma_{\text{max}}$).}

\begin{figure}[t]
    \centering
    \includegraphics[width=0.7\linewidth]{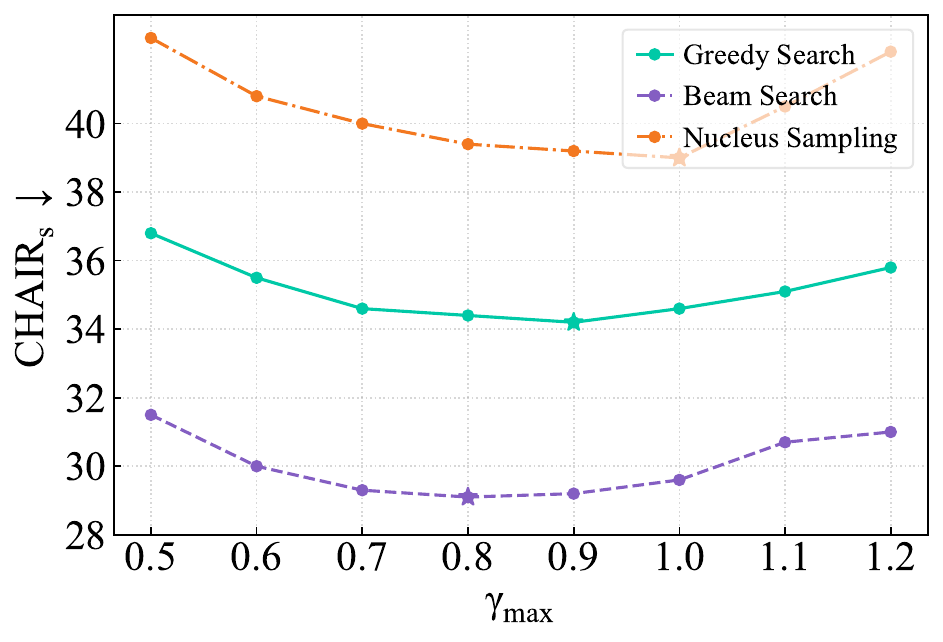}
    \caption{Hyperparameter study on the maximum suppression strength
($\gamma_{\text{max}}$). The CHAIR$_\text{S}$ score (lower is
better) is shown as a function of $\gamma_{\text{max}}$, validating
$\mathbf{\gamma_{\text{max}}=0.9}$ as the optimal trade-off point.}
    \label{fig:gamma}
\end{figure}

As shown in Figure~\ref{fig:gamma}, the $\text{CHAIR}_\text{S}$
score across all decoding strategies performs best when
$\gamma_{\text{max}}$ is in the upper-middle range. The curve shows
a broad region of strong performance, indicating that the spectral
modulation mechanism is robust to parameter choices. High
$\gamma_{\text{max}}$ values eventually lead to performance
saturation and decline, which confirms that over-suppression harms
coherence by removing necessary spectral components. We select
$\mathbf{\gamma_{\text{max}}}$ at the optimal trade-off point to
maximize hallucination mitigation while preserving essential
information.

\paragraph{Effect of Logit Fusion Weight ($\beta$).}

\begin{figure}[t]
    \centering
    \includegraphics[width=0.7\linewidth]{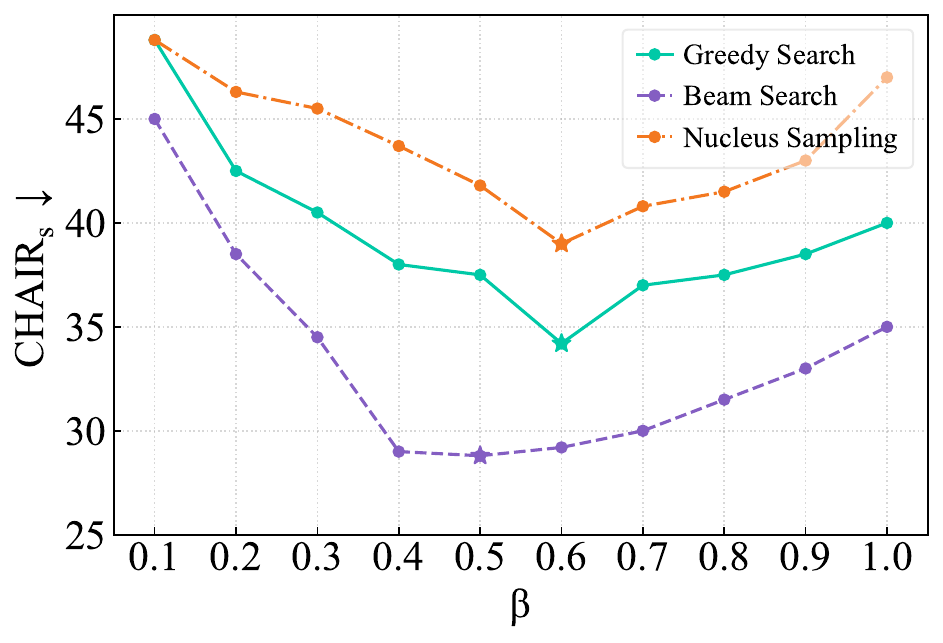}
    \caption{Hyperparameter study on the soft logit fusion weight ($\beta$).}
    \label{fig:beta}
\end{figure}

The weight $\beta$ controls the trade-off between the raw
information from the final layer and the stabilized prediction
derived from the cross-layer anchor. As illustrated in
Figure~\ref{fig:beta}, performance exhibits a clear concave trend,
which validates the necessity of a fusion mechanism:
\begin{enumerate}
\item Optimal performance is consistently achieved when $\mathbf{\beta}$
is set to the middle range, which confirms that a balanced
contribution from both the final layer and the stable anchor is
critical for maximum correction.
\item Performance begins to decline when $\beta$ deviates significantly
from the optimal range. This demonstrates that either insufficient
stability at low $\beta$ or over-reliance on the anchor at high
$\beta$ reduces the overall prediction quality.
\end{enumerate}
Based on this consistent peak, we set the soft logit fusion weight
to $\mathbf{\beta=0.6}$.

\subsection{Ablation Study}

To rigorously evaluate the contribution of each core mechanism
within the LISA framework, we perform an ablation study using
Qwen2.5-VL as our testbed. LISA comprises three main innovations:
Spectral Modulation (SM), Cross-layer Fusion (CF), and Token-wise
Soft Fusion (TSF). The results, summarized in
Table~\ref{tab:ablation_study}, validate that the combination of
these layer-wise controls is essential for maximizing hallucination
reduction and ensuring robustness across various decoding settings.

\begin{table}[t] \footnotesize
\centering \resizebox{1.0\columnwidth}{!}{
\begin{tabular}{lcccccc}
\hline
\textbf{Variant} & \textbf{SM} & \textbf{CF} & \textbf{TSF} & \textbf{Greedy} & \textbf{Beam} & \textbf{Nucleus} \\
\hline
\textbf{LISA} & \cmark & \cmark & \cmark & \textbf{85.04} & \textbf{87.09} & \textbf{85.36} \\
\hline
LISA w/o SM & \xmark & \cmark & \cmark & 84.11 & 83.81 & 84.24 \\
LISA w/o CF & \cmark & \xmark & \cmark & 84.67 & 85.54 & 84.83 \\
LISA w/o TSF & \cmark & \cmark & \xmark & 84.87 & 86.52 & 85.19 \\
\hline
Vanilla & \xmark & \xmark & \xmark & 83.93 & 84.20 & 84.10 \\
\hline
\end{tabular}}
\caption{Ablation study of LISA's core components on
\textbf{Qwen2.5-VL}. The table reports the average POPE F1 score
($\uparrow$) across three decoding strategies (Greedy, Beam,
Nucleus), validating that the layered combination of \textbf{SM}
(Spectral Modulation), \textbf{CF} (Cross-layer Fusion), and
\textbf{TSF} (Token-wise Soft Fusion) is essential for maximizing
hallucination mitigation and grounding.} \label{tab:ablation_study}
\end{table}

\paragraph{Experimental Variants.}

We compare the full LISA framework with the following essential
variants:
\begin{enumerate}
\item \textbf{LISA} (full): The complete framework, including Spectral
Modulation (SM), Cross-layer Fusion (CF), and Token-wise Soft Fusion
(TSF).
\item \textbf{LISA w/o SM} (\textit{disabling stabilization}): We disable
the \textit{Spectral Modulation (SM)} mechanism by setting
$\gamma_{\text{max}} = 0.0$. This variant performs fusion and
selection based purely on the raw spectral energy of the unstable
representations.
\item \textbf{LISA w/o CF} (\textit{disabling aggregation}): This variant
disables \textit{Cross-layer Fusion (CF)}. Specifically, we prevent
the creation and use of the stable fused representation
$\mathbf{H}_{\text{LISA}}$, forcing the subsequent TSF module to
select only from individual layer representations.
\item \textbf{LISA w/o TSF} (\textit{disabling adaptive routing}): We
perform SM and compute the stable representation
$\mathbf{H}_{\text{LISA}}$, but bypass the final \textit{Token-wise
Soft Fusion (TSF)} step. Instead, the model relies only on the final
layer logits ($H^{(L)}$).
\end{enumerate}

\paragraph{Results and Analysis.}

As shown in Table~\ref{tab:ablation_study}, the complete LISA
framework consistently achieves the best average POPE F1 score
across all tested decoding strategies, confirming the effectiveness
of integrating its three controls. The ablation results validate the
necessity and layer-wise function of each component:
\begin{enumerate}
    \item Removing \textit{Spectral Modulation (SM)} causes the most severe
performance degradation, sometimes falling below the vanilla
baseline. This highlights SM as the indispensable first step for
stabilizing the model.
    \item Excluding \textit{Cross-layer Fusion (CF)} results in a notable
decline. CF is necessary for robust multi-layer integration, as it
aggregates stabilized layers into a reliable anchor
($\mathbf{H}_{\text{LISA}}$).
    \item Removing \textit{Token-wise Soft Fusion (TSF)} causes a modest yet
consistent drop. TSF acts as the final adaptive routing mechanism,
ensuring the stable representation is utilized effectively,
especially under high-risk decoding strategies.
\end{enumerate}
In summary, the results confirm that LISA functions as a layered
system: SM prevents destabilization, CF enhances robustness through
multi-layer aggregation, and TSF adaptively exploits this stability
during decoding. Together, these components yield consistent
improvements over the vanilla baseline.

\section{Conclusion}

We present LISA, a training-free decoding approach that mitigates
hallucinations in MLLMs through \textbf{layer-wise integration and
suppression}. Our deep analysis of token trajectories and spectral
energy patterns uncovers functional distinctions across layers,
motivating a layer-wise decoding design that preserves stable
signals, facilitates semantic fusion, and suppresses unstable
activations.

LISA introduces no additional training cost and is broadly
applicable to existing MLLMs. Extensive experiments across multiple
benchmarks validate its effectiveness in reducing hallucinations and
enhancing output fidelity. These findings highlight the potential of
layer-aware decoding and open new avenues for dynamic layer
selection, adaptive routing, and post-hoc consistency verification.

\bibliography{aaai2026}

\end{document}